\pdfoutput=1
\documentclass[11pt]{article}

\usepackage{naacl2021}

\usepackage{times}
\usepackage{latexsym}
\usepackage{amsmath}
\usepackage{tabularx}
\usepackage{graphicx}
\usepackage{xcolor}
\usepackage{booktabs}
\usepackage{amssymb}
\usepackage{makecell}
\usepackage{subfigure}
\usepackage{multirow}
\usepackage{enumitem}
\usepackage{afterpage}
\usepackage[T1]{fontenc}
\usepackage[utf8]{inputenc}

\usepackage{microtype}
\newcommand{\datasetname}{\textsc{MultiOpEd}}
\newcommand{\website}{\textsc{ThePerspective}\ }

\title{\vspace*{-0.5in}

{%{\small \hfill NAACL'21}\\
\vspace*{.25in}} \datasetname: \\A Corpus of Multi-Perspective News Editorials}

\author{
  Siyi Liu \hspace{1cm} Sihao Chen \hspace{1cm} Xander Uyttendaele \hspace{1cm} Dan Roth\\
  University of Pennsylvania \\
  \small \texttt{\{siyiliu, sihaoc, xanderu, danroth\}@seas.upenn.edu} }

\date{}

\begin{document}

\maketitle

\begin{abstract}
We propose \datasetname \footnote{The authors would like to thank Daniel Ravner, the CEO of \url{www.theperspective.com}, for granting access to data from the site for academic research. }, an open-domain news editorial corpus that supports various tasks pertaining to the argumentation structure in news editorials, focusing on \emph{automatic perspective discovery}. 
News editorial is a genre of persuasive text, where the argumentation structure is usually \emph{implicit}. However, the arguments presented in an editorial typically center around a concise, focused thesis, which we refer to as their \emph{perspective}. \datasetname\ aims at supporting the study of multiple tasks relevant to  \emph{automatic perspective discovery}, where a system is expected to produce a single-sentence thesis statement summarizing the arguments presented. We argue that identifying and abstracting such natural language \emph{perspectives} from editorials is a crucial step toward studying the implicit argumentation structure in news editorials. 

We first discuss the challenges and define a few conceptual tasks towards our goal.
To demonstrate the utility of \datasetname\ and the induced tasks, we study the problem of perspective summarization in a multi-task learning setting, as a case study. We show that, with the induced tasks as auxiliary tasks, we can improve the quality of the perspective summary generated. We hope that \datasetname\ will be a useful resource for future studies on argumentation in the news editorial domain.

\end{abstract}

\section{Introduction}

% \siyi{Paper TODOs check list}
% \siyi{\begin{enumerate}
%     \item \textbf{ More comparison to other datasets using two tables} 
%     \item figures/subfigures showing examples of induced tasks \checkmark
%     \item \textbf{a better table and presentation for induced tasks} \checkmark
%     \item more dataset statistics
%     \item crowdsource results
%     \item final results numbers \checkmark
%     \item a table showing examples of how our auxiliary tasks help \checkmark
%     \item merging experiments/evaluation/results \checkmark
% \end{enumerate}}

\emph{News editorial} is a form of persuasive text that conveys consensus opinion on a \emph{controversial topic} from the editors of a newspaper. 
Much like an argumentative essay, a news editorial centers around a thesis, which represents the authors' \emph{perspective} on the topic. Usually, a news editorial argues in favor of the authors' \emph{stance} on the topic, and is substantiated by extensive factual \emph{evidence}. As news editorials function as professionally produced written discourse for conveying media attitude and guidance, they have traditionally been studied by the community as a rich resource for many argumantation-related tasks. \cite{wilson-wiebe-2003-annotating, yu2003towards, bal2009towards}.  

% Researchers have investigated and developed methods on news articles classification, summarization, and even news articles writing tasks. 
% However, existing work and datasets have not addressed the argumentative perspectives in \emph{News Editorials} enough.   
\begin{figure}
    \centering
    \includegraphics[width=7.8cm]{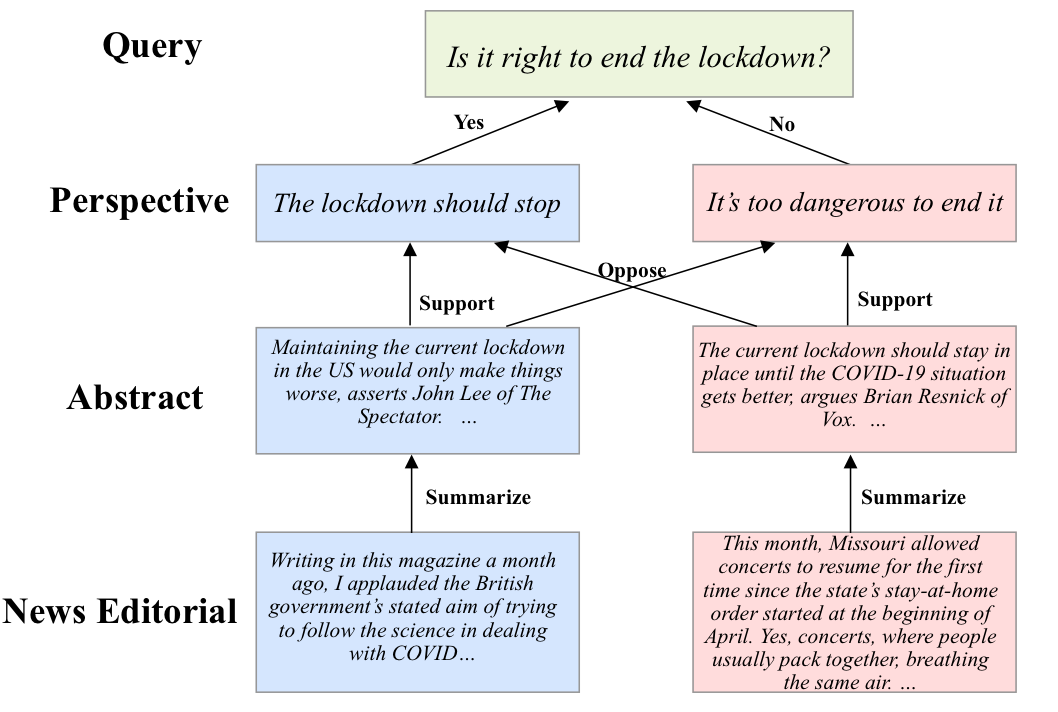}
    \caption{Structure of \datasetname . For each \emph{query} on a controversial topic, two (rather long) news editorials respond to the query from different point-of-views. Each editorial comes with a single paragraph \emph{abstract} plus a one-sentence \emph{perspective}, that abstractively summarizes the editorial's key argument in the context of the query. The two resulting perspectives serve as responses with opposite \textit{stance} to the query.}
    \label{fig:dataSample}
\end{figure}

% Arguments in \emph{news editorials} imply the authors' perspective on a news topic. It constructs as a good summarization of the author's view. 

%Arguments are the key components when people debate on different topics and express their perspectives on them. 
% Professionals are capable of summarizing from their supporting evidences to generate thesis arguments that concisely state their opinions. 
% However, arguments summarization is a challenging task for machines as it involves both natural language understanding and generation. Recent advances on summarization system \citep{lewis-etal-2020-bart} benefited from large amounts of pre-trained data have been proved successful, but they did not address the problem of summarizing a document to an argumentative sentence enough. The aim of arguments summarization is to produce a concise argumentative sentence given sentences (paragraph or article) discussing on a topic. Besides basic expectations of a strong summarization system, where a summarized sentence should be concise and meaningful, a successful argumentative summarization system should generate a sentence consist of the following properties: 1. the sentence is argumentative; 2. it is relevant to the news topic discussed; 3. it maintains the same stance on the topic with the paragraph it is summarized from.

This work targets
the problem of developing computational methods to identify and comparatively analyze the authors' \emph{perspectives} and supporting arguments behind news editorials. 
%However, 
One challenge to studying the argumentation structure in news editorials is that its elements are rarely expressed explicitly \cite{el-baff-etal-2018-challenge}. For example, Figure~\ref{fig:dataSample} shows two news editorials holding opposite views on whether a lockdown should continue. However, neither of them 
present their key perspectives explicitly. %in natural language. 
%
% Instead, more subtle and implicit rhetoric strategies are employed to convey the author's opinion \cite{van1995opinions}. As a result, the comprehension of editorials usually requires reasoning over extensive, contextual knowledge \cite{reed-etal-2008-language}. 
Instead,  the \emph{perspective} is %usually 
conveyed through subtle rhetoric strategies to either affirm or challenge the readers' stance from prior belief on the topic, as a study by \citet{el-baff-etal-2018-challenge} discovers. As %the example in
Figure~\ref{fig:dataSample} shows, the statement ``\emph{The lock down should stop}'' concisely summarizes the perspective expressed in the article on the left. We refer to such statements as ``\emph{perspectives}'' throughout the paper. 
The ability to abstractively summarize the perspectives from the editorial would allow us to understand multiple topic-aligned editorials in context and reason about their inter-editorial argumentation structure.

To facilitate research along the line, we collect data from \textsc{ThePerspective}\footnote{\url{https://www.theperspective.com/perspectives/}} website, and construct \datasetname, an open-domain English news editorial corpus that supports various tasks pertaining to the argumentation structure in news editorials, focusing on \emph{automatic perspective discovery}  \cite{CKYCR19}. 
The structure of the data is shown in Figure~\ref{fig:dataSample}. For each of the 1,397 natural language \emph{query} on a different topic in our dataset, it features two (rather long) news editorials.  Each editorial features a single-sentence perspective, which is abstractively summarized from the editorial by human experts. A short \emph{abstract} then highlights the details in the editorial that support the
%provides the highlight of supporting details in the editorial to the
\emph{perspective}. 
The \emph{perspectives} of the two editorials represents responses of opposite \emph{stances} towards the query.

Naturally, the structure of the dataset induces a range of important %various 
argumentation-related natural language understanding tasks. For instance, the presence of the summary \emph{perspective} allows for stance classification \cite{hasan-ng-2013-stance} with respect to the query, which arguably is more tangible than inferring the stance from the entire editorial. Another example task is the conditional generation of the \emph{perspective} from the abstract/editorial, which relates to the widely studied task of \emph{argument generation} \cite{hua-wang-2018-neural, alshomary-etal-2020-target}. We defer the more detailed description of the induced tasks to Section \ref{sec:induced_tasks}. 
% Given the diverse attributes provided by the \emph{Perspective} dataset, it induces different tasks involved argumentations, for instance, stance and relevance classification. We construct a total of ten different potential tasks on this dataset and discuss them in Section \ref{sec:3}.
%or instance, argument quality, stance and relevance detection from pairs among one-sentence perspectives, multi-sentence summaries, and longer articles, and summary generations from multi-sentence summaries and articles. These varied tasks in one dataset provides us intuition to implement multi-task setting for one target task.

% To generate an argumentative summary that is both relevant to the topic and implies consistent stance, we propose a multi-task learning setting for our argument summarization system. We finetune SOTA summarization system, BART, and add two auxiliary classification tasks for our target argument summarization task: relevance and stance classification. Our results show that the auxiliary signals from them effectively help the generation results. 
One key advantage of \datasetname\ that is absent from earlier datasets is that a large number of argumentation-related tasks can be studied jointly using a single high quality corpus. To demonstrate this benefit and the utility of the \datasetname\ dataset \footnote{Our code and data is available at \url{http://cogcomp.org/page/publication_view/935}}  along with its induced tasks, we study the problem of perspective summarization in a multi-task learning setting. We employ perspective relevance and stance classifications as two auxilliary tasks to the summarization objective. Our empirical and human analysis on the generated summaries show that the multi-task learning setting improves the generated perspectives in terms of the argument quality and stance consistency.

% TODO: how to address the point that our paper can do so many tasks
In summary, our contributions in this work are three-fold. First, we propose a conceptual framework for identifying and abstracting the \emph{perspectives} and the corresponding argumentation structure in news editorials, and define a set of tasks necessary for %us to achieve
achieving this goal. Second, we propose the \datasetname\ dataset, a news editorial dataset that induces multiple %various 
argumentation-related tasks. Third, we demonstrate the utility of our multi-purpose dataset and induced tasks, by using the perspective summarization task as a case study. We include the induced tasks as auxiliary objectives in multi-task learning setting, and demonstrate their effectiveness to  perspective summarization. 
% \begin{itemize}
%   \item an argumentation dataset in news domain that induces different generation and classification tasks
%   \item a multi-task learning model for argument summarization that regularizes the learning process using stance and relevance auxiliary classification tasks
% \end{itemize}

% TODO:
% how should we address the point that this multitask setting work even when we do classification to help generation

% how should the articles play in as we never used them

\section{Design Principles} \label{sec:design_principle}
\begin{table*}[t]
    \centering
    \begin{tabular}{l|cccc}
         \textbf{Dataset} & \textbf{Source} & \textbf{Open Domain} & \textbf{Cross Article} & \textbf{Abstractive}\\
        \hline
         \hline
         \textsc{AraucariaDB}~\cite{reed-etal-2008-language} & News Ed. & \checkmark & $\times$ & $\times$\\
         \cite{stab-gurevych-2014-annotating} & Essay & \checkmark & $\times$ & $\times$ \\
         \cite{eckle-kohler-etal-2015-role} & News & \checkmark & $\times$ & $\times$ \\
         \cite{hua-wang-2018-neural} & Reddit/Wiki. &  \textit{Politics} only & $\times$ & $\checkmark$\\
         \textsc{Perspectrum}~\cite{CKYCR19} & Debate & \checkmark & $\times$ & $\checkmark$\\
         \hline
         \datasetname & News Ed. & \checkmark & \checkmark & \checkmark \\
         \hline
    \end{tabular}
    \caption{A comparison across datasets with similar purpose to \datasetname. We compare the datasets along three dimensions. \textit{Open Domain}: whether the dataset features a wide variety of topics.  \textit{Cross Article}: whether the argumentation structure between documents are annotated. \textit{Abstractive}: whether the elements in argumentation structure is abstractive or extractive. }
    \label{tab:datasets_comparison}
\end{table*}

% \dr{More importantly, this section is a bit weak, and it feels like it ends in the middles. My understanding was that the goal here is to have a focused related work discussion and use it to distinguish the design principles we have from other work; this is done only partially, since it does not cover (i) the importance of the multi-purpose dataset, (ii) the availability of short and long pieces of text that provide opportunities other dataset do not have, (iii) the paired structures that, like-wise, provide opportunities other datasets do not have, and (iv) the existence of a trigger query, that most other datasets do not have (and there are probably other issues).}

Our goal of perspective discovery follows similar definition proposed by \citet{CKYCR19}, and is closely related to a widely studied area of argumentation mining, i.e. identifying the argumentation structure within persuasive text   \cite{stab-gurevych-2014-identifying, kiesel-etal-2015-shared}. 
However, most studies in this domain focus on extractive methods, which becomes less applicable to our study. As the arguments are usually presented in an subtle and implicit way in news editorials, we instead focus on the generation methods for the perspectives. This closely resembles the argument conclusion generation task \cite{alshomary-etal-2020-target}. One key distinction here is the presense of \emph{query} to provide topic guidance during the perspective generation.

Compared to other conditional text generation tasks, perspective generation subjects to a few more constraints with respect to the argumentation structure. For example, the perspective must constitute the same stance \cite{hasan-ng-2013-stance} as the editorial towards the query. On the other hand, while the editorial may cover content not directly related to the query, the generated perspective must present a relevant argument in the query's context. Such structural constraints can be studied in the format of classification problems. And being able to study such problems along side the perspective summarization task on one high-quality corpus is important in our case, as it opens up the probability of modeling the tasks jointly. We show the benefit of doing so by presenting a case study in section \ref{sec:multi_task}.

% As such, structure is usually not presented in surface forms in editorials, and current extractive methods for argumentation mining become less applicable to our study. 

% \dr{I am not sure what you mean when you say "structure" here; also, could we say that the setting here is more realistic?} 
As the query provides topic guidance, it allows for the study of the topic-aligned pairs of editorials which presents counter-arguments to each other. Such property is absent from notable datasets of similar purposes to ours, as shown in Table~\ref{tab:datasets_comparison}. \textsc{AraucariaDB} \cite{reed-etal-2008-language} is the first effort to provide large-scale annotations of dense argumentation structure within individual news editorials. \citet{stab-gurevych-2014-annotating, eckle-kohler-etal-2015-role} provide resources for %on
extractive argumentation structure in persuasive essays and news articles, respectively. Later works \cite{hua-wang-2018-neural,CKYCR19} 
focus on the abstractive generation or identification of arguments from web corpora. All of these datasets focus on studies of argumentation structure within individual document.
Instead, our proposed dataset presents the opportunity to study the cross-document argumentation structure. 

\begin{table}[h]
    \centering
    \begin{tabular}{lcccc}
        \hline
         Instance & Size & Avg. Len. & Min & Max
         \\
         \hline
         \hline
         \textit{Query} & 1,397 & 7.4 &3 &15 \\
         \textit{Perspective} & 2,794 &6.1 & 2 &10\\
         \textit{Abstract} & 2,794 & 101.9 & 47 &160\\
         \textit{Article} & 2,584 & 918.6 & 74 & 7,608 \\
         \hline
    \end{tabular}
    \caption{Statistics of the \datasetname\ dataset. Size represents the number of each valid instance, Avg. Len. indicates the average length of each instance in terms of the number of tokens split by space, and Min and Max represent the number of tokens of the shortest and longest texts of each instance.}
    \label{tab:dataStats}
\end{table}

\section{\datasetname\ and Induced Tasks} \label{sec:induced_tasks}
Following the design principles outlined in the previous section, we propose a topic-aligned English news editorial corpus, \datasetname. The structure of an example instance in \datasetname\ is shown in Figure~\ref{fig:dataSample}. To clarify our description of the dataset, we use the following notations.  Let $q$ be a query about a controversial topic. Each $q$ in the dataset is paired with two editorials $e_{pro}$ and $e_{con}$, that constitute \textit{supporting} and \textit{opposing} stances to the query $q$ respectively. Each editorial is abstracted into and a single-sentence \emph{perspective} $p$, which provides a high-level summarization of the key argument presented in the editorial. The premises, or relevant details to support the perspective, forms the abstract $a$. 

Naturally, the relation between these elements induces several tasks, most of which encompass similar definitions to existing argumentation-related tasks. We define and describe the tasks and their connection to our end goal of perspective discovery below. 

\begin{enumerate}[leftmargin=*]
    \item \textit{Generating an Abstract}: Given an editorial $e$, a system is expected to identify and summarize the relevant arguments into an abstract paragraph $a$ to the context provided by the query $q$. This is closely related to the task of argument synthesis \cite{el-baff-etal-2019-computational, hua-etal-2019-argument-generation}. We set aside this problem in our case study in section \ref{sec:multi_task}, and use the abstract provided by the dataset. 
    
    \item \textit{Perspective Summarization}: Given the generated abstract $a$ and the query $q$, a system is expected to generate the perspective $p$, a concise summary of the arguments presented in $a$. Conceptually, this problem resembles the task of argument conclusion generation \cite{alshomary-etal-2020-target}. We adopt a slightly different setting where the target topic is expressed in the form of a natural language query. 
    
    \begin{figure}[h]
    \centering
        \includegraphics[width=5cm]{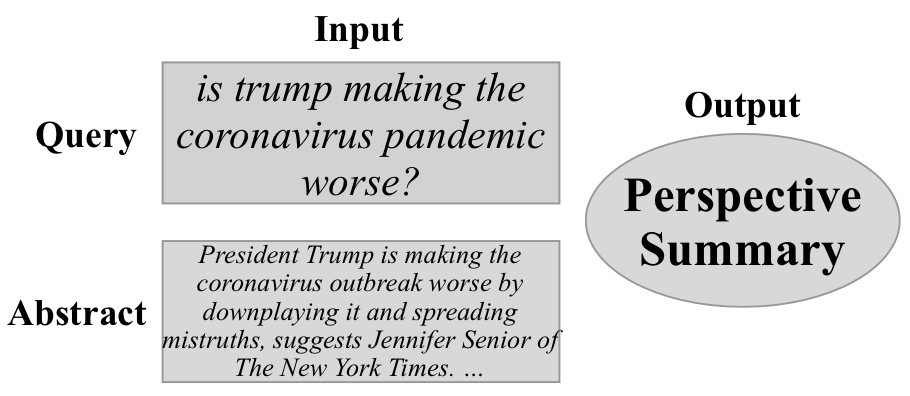}
        \caption{Perspective Summarization: Generate an single sentence argument that represents the key \emph{perspective} expressed in the news editorial.}
    \end{figure}

    \item \textit{Stance Classification}: Our goal is to infer the editorials $e$'s stance towards a query $q$. The generated perspective $p$ from editorial $e$ allows us to focus on a simpler task definition of classifying the stance of the perspective to the query $q$ \cite{hasan-ng-2013-stance, bar-haim-etal-2017-stance}.
        
    \begin{figure}[h]
    \centering
        \includegraphics[width=5cm]{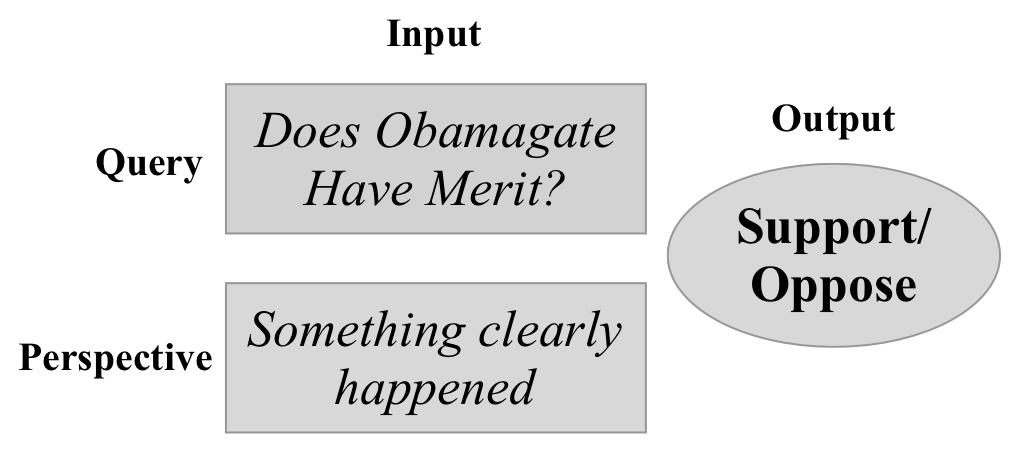}
        \caption{ Stance Classification: Decide if the perspective supports or opposes the query.}
    \end{figure}
    
    \item \textit{Assessing the Relevance of Perspective}: We want to measure the validity of the perspective by assessing whether the perspective presents a relevant argument towards the query \cite{CKYCR19, ein2020corpus}. This can be formulated as a classification problem with the query $q$ and a perspective $p$ as inputs, as we show in section \ref{sec:multi_task}.

    \begin{figure}[h]
    \centering
        \includegraphics[width =5cm]{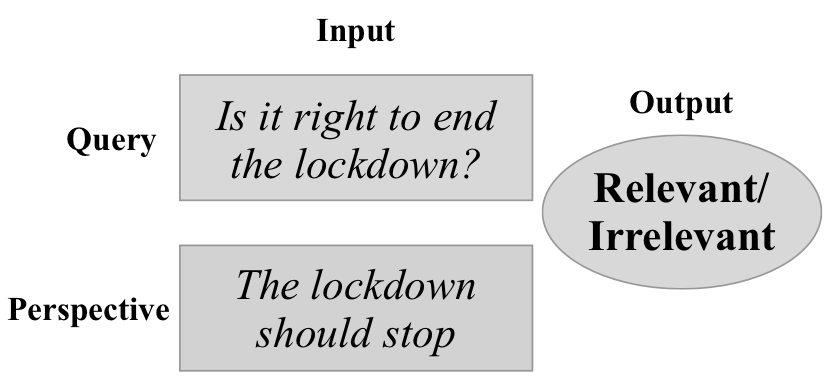}
        \caption{ Relevance classification: Decide if the perspective is relevant to the query.}
    \end{figure}

\end{enumerate}

\section{Dataset Construction}
\subsection{Data Collection}
% We collected the data by crawling from the
We extract the query, editorial article pairs, abstract paragraph pairs, along with their perspective summaries from
\textsc{ThePerspective}\footnote{\url{https://www.theperspective.com/perspectives/}} website. The website presents controversial topics in the form of queries. For each query, two related editorial articles with opposing views from different sources are selected by the writers from the website. The writers create a concise one-sentence summary of each article as the response to the query, and an abstract paragraph to summarize the relevant arguments from the article. An example structure of the data is shown in Figure.\ref{fig:dataSample}.
% The editorials are summarized by human experts into a perspective, accompanied by a short paragraph of supporting details. We collect 1,397 queries from the website, keep the structure of data as given, and annotated it in the next step. 

We use \textsc{BeautifulSoup} \footnote{\url{https://www.crummy.com/software/BeautifulSoup/bs4/doc/}} and \textsc{Newspaper3k} \footnote{\url{https://newspaper.readthedocs.io/}} to extract and clean the perspective and news data.

\subsection{Crowdsource Verification \& Annotation}
\label{sec:4_2}
To verify the structure from the website, and collect additional annotations, such as stance of the perspectives, we conduct a few annotation experiments with Amazon Mechanical Turk\footnote{\url{https://www.mturk.com/}}. 
% We use Amazon Mechanical Turk\footnote{https://www.mturk.com/} to verify the structure of our corpus and collect additional annotations for missing properties in the dataset. 
% For all of our annotations, we require the Turkers to qualify as Masters, Turkers that are identified by Mechanical Turk as high performing workers and continue to pass their statistical monitoring, and locate in the United States. 
For all of our annotation experiments, we require the workers to be located in the United States, as the controversial topics covered by the website are most applicable in the U.S. context. 
We also require the workers to have masters qualifications (i.e. Top performers recognized by MTurk among all workers). 
We compensate the workers $\$0.75, \$1.00$ and $\$1.25$ per 10 queries for the implicit reference resolution, topic annotation, and stance annotation tasks respectively. The compensation rates are determined by estimating the average completion time for each annotation experiments.
% according to the federal minimum wage and our approximate of time needed for each batch. 
Example screenshots of our annotation interface and more detailed annotation guidelines can be found in Appendix \ref{sec: screenshots}.
% Besides, for some difficult annotation tasks, we asked a linguistic expert to annotate the data and verified the annotations with crowdsource, given that crowdsourced annotations for challenging tasks may not be reliable.

\subsubsection{Stance Annotation}

%TODO How to phrase it here
In our dataset, each \textit{query} is presented with two \textit{perspectives} with opposite stance to the query. 
However, the raw data that we collected does not specify the stance of each perspective individually. 

% However, the raw data that we collected does not specify which argument supports the query, and which disagrees with it. 
% To augment the dataset,
We ask two expert annotators label whether each perspective is offering a supporting or opposing view with respect to its query. The two experts discuss and adjudicate their decisions. We then ask on average three crowdsource workers per instance to verify the annotations. 
% During the annotation process of our domain expert, 

From the annotations collected by experts, we find that $30$ out of $1,397$ queries do not constitute a clear stance. Such queries are typically "open-ended" questions which cannot be responded with a \texttt{yes} or \texttt{no} answer, i.e. \texttt{why} or \texttt{what} questions. We leave these instances unlabeled and exclude them from the next verification step. 
% 30 out of 1397 queries were found to be\textit{open-ended}, meaning that the query can not be easily answered by "Yes" or "No". 

% These queries  generally ask for opinions or reasons that can not be generalized to only two stances. For instance, \textit{"Why did Trump hire Scaramucci?"} is a query that asks for the reasons behind an event. There are numerous potential answers to this question and they can not be generilizaed to either "support" or "oppose". 

% For the rest of the dataset, our domain expert annotated the stance a perspective takes on its associated query, either \textit{support} or \textit{oppose}. For instance, for a query \textit{"should we end covid-19 security measures soon?"}, the perspective \textit{"We need to protect our economy"} is labeled as "support" and \textit{"The lockdown is still necessary"} is labeled as "oppose". 

To assess the quality of stance labels created, we randomly sample 500 perspectives, and ask three MTurk workers per instance to verify stance labels.
% We then randomly sample 500 (perspective, query) pairs and ask three crowdsource workers per instance to verify stance labels. 
We computed the inter-rater agreement fleiss' $\kappa = 0.81$ among workers, and the agreement between majority decision from works and the expert's adjudicated annotations is cohen's $\kappa = 0.92$. We describe how we measure the two types of agreements respectively in Appendix \ref{sec:agreement}. 
\subsubsection{Implicit Reference Resolution in Perspectives}
Some of the perspectives in our dataset have implicit references to certain subjects in the query. For instance, for a query \textit{``Is Trump Right To Criticize Mail-In Voting?''}, and a perspective \textit{``It's far too risky for an election''}, the word "It" in the perspective refers to ``Mail-in Voting'' in the query. 
% If a system is only given inputs as perspectives sentences, without the query, 
% this kind of perspectives that are not stand-alone valid may not be a good enough sets of data, since  reading at the perspectives alone, the system, or even humans, will not be able to infer what is the key subject being discussed and what is the "It" referring to. 
As we assume that a perspective should presents a complete, valid argument on itself, we decide to replace such implicit reference in a perspective with the correct referent in the query. For example, the corrected perspective in the previous example would become \textit{``Mail-in voting is far too risky for an election''}.

%If we simply use this perspective as an input of the model, it may suffer from a problem that it does not know which subject the "It" in the perspective refers to, since we do not have a structure in the model that provides this information. So the original perspective sentence may not be stand-alone valid as an argument that contains background information of the subjects discussed.

% To alleviate this problem and make our dataset more robust to potential tasks, 
We ask one expert annotator to identify implicit references and make modifications for every perspective in the dataset.
% we asked a domain expert to modify the perspectives by replacing the implicit references to the query with their referents, 
% i.e., the subjects referred to in the query. For instance, for the perspective \textit{"It's far too risky for an election"}, it will be modified to \textit{"Mail-in voting is far too risky for an election"}, by replacing the reference "it" in the original perspective with its referent "Mail-in voting" from the query. 
In total, $1,301$ out of $2,794$ perspectives are identified and corrected by the expert annotator. 
We ask three Turkers to verify that the modifications do not introduce any grammatical error or change the original meaning. We randomly sample 500 modified perspectives and present Turkers with the question of "Will this modification change the original meaning or introduce grammar error?". 
The percentage of majority answers being ``No'' is 84\%. 
We include both changed and original versions of the perspectives in our datasets. 
% original perspectives and perspectives with modifications aimed to support future research depending on different usage of perspectives in our dataset.

%To solve this problem, we aimed to preprocess it by replacing the implicit references to the query with their referents. For the above example, we want to modify the perspective argument to "Mail-in voting is far too risky for an election", so it contains every information we need to understand the subjects. We first tried to use a SOTA coreference system to help generate better perspective arguments; however, some of the examples are challenging, and current coreference system cannot provide reliable enough labeling for us. So we manually modified the perspective sentence by replacing the implicit references to the query with their referents. We asked the crowdsource workers to rate the quality of our modification and calculated the agreement. In our experiments, we use the perspectives after this modification as inputs to the model. 

\subsubsection{Topic Annotations}

\begin{figure}
    \centering
    \includegraphics[width=0.95\linewidth]{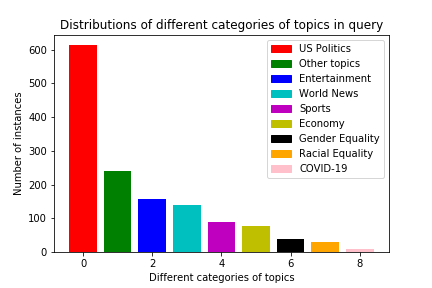}
    \caption{Topic distribution of the $1397$ \emph{queries} in \datasetname. Note that the two editorials for each query fall in the same topic category as the query.}
    \label{fig:topic_distributions}
\end{figure}

% The $1,397$ queries in our dataset feature 1,397 distinct popular news topics or events. However, some of them are in the same category of topics and could share similar potentially useful linguistic attributes. Therefore,
We create 9 topic labels according to the categorization from \website website and major news outlets. We then as ask three MTurk workers  to assign one of the 9 topic labels to each query. We regard the majority answer by the Turkers as the annotation for its topic category. In cases where all three annotators choose different categories ($43$ cases out of all $1397$ queries), we label it as \texttt{other topics}. We show the distribution of topic categories in Figure \ref{fig:topic_distributions}. The inter-agreement among three annotators for this 9-class classification task is $\kappa = 0.65$

% SC: move this to appendix, or delete this
% \begin{table}
%     \centering
%     \begin{tabular}{l|c}
%         \hline
%          \textbf{Categories of topics} & \textbf{\# of queries}  \\
%          \hline
%          \textit{US Politics} & 614  \\
%          \textit{Entertainment} &  158 \\
%          \textit{World News} & 139 \\
%          \textit{Sports} & 90  \\
%          \textit{Economy} & 78  \\
%          \textit{Gender Equality} & 38  \\
%          \textit{Racial Equality} & 29  \\
%          \textit{COVID-19} & 10  \\
%          \hline
%          \textbf{Total} & 1397\\
%          \hline
%     \end{tabular}
%     \caption{Query distributions on different categories of topics it discusses on}
%     \label{tab:data_distri}
% \end{table}

\subsection{Dataset Statistics}
\datasetname \ consists of 1,397 queries about different news topics. Each query is presented with two perspectives, two abstracts and two linked news editorials. 
 Despite a few stale urls and invalid re-directions, we manage to extract the text for 2,584 news editorials. More detailed statistics are reported in Table \ref{tab:dataStats}.

\begin{table*}[t]
\begin{tabular} {c|cccc|cc}
\hline
Model & \textsc{Rouge}$_1$ & \textsc{Rouge}$_2$ & \textsc{Rouge}$_L$ & \textsc{BertScore} & \textsc{Rel.} \% & \textsc{Stance} \%\\
\hline
\textsc{Bart} & 28.24 & 11.34 & 26.96 & 88.67 & 91.91 & 72.32 \\
\textit{+ Rel} & 28.35& 11.51& 27.12 & 88.69& 92.98 & 72.68 \\
\textit{+ Stance} & 28.19 & 11.53 & 26.93 & \textbf{88.75} &91.25  & 73.39 \\
\textit{+ Rel $\&$ Stance}  & \textbf{29.18} & \textbf{11.92} & \textbf{27.94} & 88.74 & \textbf{94.64}& \textbf{74.29} \\
\hline
\end{tabular}
\centering
\caption{Results of our multitask \emph{perspective} summarization models. We compare to \textsc{Bart} as a baseline, and experiment with different combinations of the auxiliary tasks. We report the $F_1$ score under \textsc{Rouge}$_{\{1,2,L\}}$ and \textsc{BertScore} metrics, as well as the percentage of summaries with the correct relevance and stance label, as predicted by our pretrained classification models respectively. See Appendix \ref{sec:training_details} for training details and hyperparameters settings.}
\label{tab:result}
\end{table*}

\begin{table}[h]
\begin{tabular} {c|cc|cc}
\hline
Model  & \textsc{Rank} & \%\textsc{1st} & \textsc{Rel.} & \textsc{Stance} \\
\hline
\textsc{Bart} & 2.09 & 49.50 & 77.00 & 70.50 \\

\textit{+ Rel} & \textbf{1.74} & \textbf{60.50} & \textbf{87.00} & 70.50 \\

\textit{+ Stance} & 1.78 & 58.50 & 83.00 & \textbf{79.50} \\

\textit{+ R $\&$ S}  & 1.76 & 59.00 & 82.00 & 69.00 \\
\hline
\end{tabular}
\centering
\caption{Human Evaluations results. ``\textsc{Rank}'' shows a model's averaged rank judged by the raters ($1=$ \texttt{best}, $4=$\texttt{worst}) ``\%\textsc{1st}'' represents the percentage of generated summaries from one model that are ranked the best. We allow ties in the ranking. \textsc{Rel.}  and \textsc{Stance} are the percentages of generated summaries that are relevant to and have the correct stance with respect to the query.}
\label{tab:human_eval}
\end{table}

\section{Case Study: Multi-task Learning for Perspective Summarization} \label{sec:multi_task}

% \begin{table*}[t]
% \begin{tabular} {c|cccc|cc}
% \hline
% Model & \textsc{Rouge}$_1$ & \textsc{Rouge}$_2$ & \textsc{Rouge}$_L$ & \textsc{BertScore} & \textsc{Rel.} \% & \textsc{Stance} \%\\
% \hline
% \textsc{Bart} & 25.21 & 8.46 & 23.70 & 88.27 & 85.64 & 72.85 \\
% \textit{+ Rel} & 25.75& 9.29& 24.44 & \textbf{88.52}& \textbf{86.37} & 71.96 \\
% \textit{+ Stance} & 24.34 & 8.50 & 23.38 & 88.32 &83.75  & 72.32 \\
% \textit{+ Rel $\&$ Stance}  & \textbf{26.30} & \textbf{9.32} & \textbf{25.07} &88.51 & 86.22& \textbf{73.81} \\
% \hline
% \end{tabular}
% \centering
% \caption{Results of our multitask \emph{perspective} summarization model. We compare to \textsc{Bart} as a baseline, and experiment with different combinations of the auxiliary tasks. We report the $F_1$ score under \textsc{Rouge}$_{\{1,2,L\}}$ and \textsc{BertScore} metrics, as well as the percentage of summaries with the correct relevance and stance label, as predicted by our pretrained classification models respectively. See Appendix \ref{sec:training_details} for training details, reproducibility and hyperparameters choosing.}
% \label{tab:result}
% \end{table*}

\subsection{Multi-Task Framework}
To demonstrate the benefits of modeling the induced tasks on the argumentation structure, we present a case study on the task of perspective summarization.
Given a query and an abstract from the related editorial, a system is expected to produce a concise and fluent summary perspective for the editorial. In addition, the generated perspective ideally should satisfy a few structural constraints with respect to the query. For instance, the generated perspective must constitute the same stance as the editorial towards the query. Also the perspective should be relevant in the context of the query. The two requirements resemble the perspective \emph{stance} and \emph{relevance} classification`` tasks defined in section~\ref{sec:induced_tasks} respectively.

Motivated by this, we study the two tasks together with perspective summarization in a multi-task learning framework. We choose BART \citep{lewis-etal-2020-bart} as our base summarization model. BART is a pretrained auto-regressive transformer \cite{vaswani2017attention} encoder-decoder model, that have been proven effective in conditional text generation and other NLP tasks. 

We start with a pretrained BART base model with $139M$ parameters, and finetune the model to output the target perspective given the query and abstract concatenated as input. In addition, we put two separate linear layers over the pooled embeddings of the last decoder layer, and predict the relevance and stance labels of the generated summary respectively. The two tasks and the perspective summarization are learned jointly, and share the underlying model parameters from BART.

One obvious challenge in the setup is that we do not have access to the ground truth stance and relevance labels for the generated summaries during training. To address this, we adopt similar strategies as in knowledge distillation \cite{knowledge_distill}. We first train two separate BERT \cite{devlin-etal-2019-bert} classifiers as the \textit{teacher} models for stance and relevance classificaiton respectively. Due to the size limit of our dataset, we pretrain both models on the \textsc{Perspectrum} dataset \cite{CKYCR19}, which contain over 7,000 instances of training data, with similar formats and definition to our $(query, perspective)$ pairs. We further fine-tune the models on our training set. When measured against our test set, the relevance and stance models achieve binary accuracy of $92\%$ and $75\%$  respectively. 

During the perspective summarization model training, we use the pre-trained BERT models for relevance and stance classification to predict labels for each generated summary. We expect the BART plus linear layers to ``mimic'' the predictions made by the two pretrained BERT models respectively. Specifically: 
% Other than the summarization target task of the BART language model, we include two auxiliary tasks, stance and relevance classification. We added two classification heads on top of BART Encoder-Decoder structure. All three objectives have the shared parameters in BART Encoder-Decoder. We use finetuned classifiers to produce silver labels for stance/relevance predictions between the generated perspective and the query. We back-propagate the sum of three losses to update the BART Encoder-Decoder, the summarization language model head, and the two classification heads.
% \siyi{Specifically, we pass each input abstract $\mathcal_{A}$ and query $\mathcal_{Q}$ through the pre-trained BART model and collect their decoder's last hidden states. We take their eos-token's embedding as their sentence representations.}
\begin{align*} 
\mathcal{H}_{\textsc{Q}} &= \textsc{EOS} (\mathcal{D}_{\textsc{BART}}(\mathcal{E}_{\textsc{BART}}(\mathcal{Q}))) \\
\mathcal{H}_{\textsc{A}} &= \textsc{EOS} ( \mathcal{D}_{\textsc{BART}}(\mathcal{E}_{\textsc{BART}}(\mathcal{A})))
\end{align*} 
% \siyi{where $\mathcal{E}_{\textsc{BART}}$ and $\mathcal{D}_{\textsc{BART}}$ are the Bart encoder and decoder, and $\textsc{EOS}$ is end of sequence token (EOS token)'s vector embedding from the last hidden states. Therefore, $\mathcal{H}_{\textsc{Q}}$ and $\mathcal{H}_{\textsc{A}}$ represent the sentence embeddings of the input query and the generated perspective, respectively.}
We feed the query and the abstract separately through the BART encoder ($\mathcal{E}_{\textsc{BART}}$) and decoder ($\mathcal{D}_{\textsc{BART}}$). We get their hidden representations $\mathcal{H}_{\textsc{Q}}$ and $\mathcal{H}_{\textsc{A}}$ as the embedding of the end-of-sentence (`</s>') token from the decoder. 
We then concatenate $\mathcal{H}_{\textsc{Q}}$ and $\mathcal{H}_{\textsc{A}}$, and feed the concatenation to the two linear layers.  Finally, a softmax layer is applied to get stance/relevance predictions $\tilde{y}_{rel}$ and $\tilde{y}_{stance}$. 
\begin{align*} 
    \tilde{y}_{stance} &= 
    \textsc{Softmax}(W_s^T [\mathcal{H}_{\textsc{Q}},  \mathcal{H}_{\textsc{A}}]) \\
    \tilde{y}_{rel} &= 
    \textsc{Softmax}(W_r^T[\mathcal{H}_{\textsc{Q}}, \mathcal{H}_{\textsc{A}}])
\end{align*} 
% These predictions represent the stance/relevance prediction given the representations of generated perspective and its query.
%Given the rich information we have on each perspective, we implemented a multi-task learning model that can better exploit the advantages of our dataset. There are two general challenges with the current argument summarization systems. One is that the generated summarization may fail to relate to the topic, and another is that it may fail to take the correct stance on a topic. To overcome these challenges, we introduce two auxiliary signals to regularize our learning process. We have two auxiliary tasks, relevance prediction and stance prediction, to assist the learning process of our target task, argument summarization. The general architecture of our model is shown in Figure \ref{fig:model}. 
%To leverage the power of transfer learning and pre-trained models, we finetune our dataset on a state-of-the-art pre-trained transformer model, BART \citep{lewis-etal-2020-bart}.  
Next, We feed the query and the generated summary to the two pretrained BERT classification models to get the soft stance and relevance labels $y_{rel}$ and $y_{stance}$. We use two mean square error (MSE) loss terms to measure the discrepancy between the BART predictions and the soft labels.
\begin{align*}
    \mathcal{L}_{\textsc{Rel}}&= \textsc{MSELoss}(y_{rel},\tilde{y}_{rel})  \\
    \mathcal{L}_{\textsc{Stance}}&= \textsc{MSELoss}(y_{stance},\tilde{y}_{stance}) 
\end{align*} 
%  We then use two finetuned BERT Classification models \citep{devlin-etal-2019-bert} to predict  stance and relevance labels for the generated perspective and input query. These two classifiers are finetuned on our dataset beforehand, and we will discuss their details more in section \ref{sec:rel_bert} and \ref{sec:stance_bert}. We take them as our silver labels here for the stance and relevance classification tasks, and compare them to $\mathcal{P}_{\textsc{Stance}}$ and $\mathcal{P}_{\textsc{Rel}}$ to compute the auxiliary losses. 
%  \siyi{where $\tilde{\mathcal{P}}_{\textsc{Rel}}$ and $\tilde{\mathcal{P}}_{\textsc{Stance}}$ are the silver labels predicted by finetuned classifiers trained beforehand.}
%   The target loss is calculated using a language model head and cross entropy loss, given the decoder's last hidden states of input Abstract $\mathcal{A}$ only. The total loss that will be optimized to update the parameters in our BART model will then be: 
We combine $\mathcal{L}_{\textsc{Rel}}$ and $\mathcal{L}_{\textsc{Stance}}$ with the summarization objective, $\mathcal{L}_{\textsc{Sum}}$, which is the negative log-likelihood loss between generated and target perspective. The auxiliary losses $\mathcal{L}_{\textsc{Rel}}$ and $\mathcal{L}_{\textsc{Stance}}$ are weighted by tunable hyperparameters $\alpha _1$ and $\alpha_2$ respectively. 
\begin{align*} 
\mathcal{L} &= \mathcal{L}_{\textsc{Sum}} + \alpha _1 \cdot \mathcal{L}_{\textsc{Rel}} + \alpha _2 \cdot \mathcal{L}_{\textsc{Stance}}
\end{align*}
% $\alpha _i$ is a hyperparameter that determines the extent to penalize the model with i-th auxiliary loss, and $Loss_{Ai}$ is the i-th auxiliary loss.

% The auxiliary loss signal represents how well the generated perspective performs on the stance/relevance classification task. The higher the probability that generated perspective takes the correct stance on the query, or the more it is relevant to the query, the less it will be penalized by this auxiliary loss. This multi-task setting regularizes the training process and forces the model to generate perspective with correct stance and more relevance to the query. \siyi{We will demonstrate how the auxiliary tasks help the generation with some examples in section \ref{sec:7}}

\subsection{Results} \label{sec:results}
\subsubsection{Automatic Evaluations}
Table~\ref{tab:result} shows our evaluation results of our multi-task model with different combinations of auxiliary tasks. The reported results are averaged over three trained models with different random initialization. We first evaluate the generated perspective summaries against the target perspective with \textsc{Rouge} \cite{lin-2004-rouge} and \textsc{BertScore} \cite{bert-score} metrics. 
We observe that relevance and stance auxiliary tasks both increase the \textsc{Rouge} and \textsc{BertScore}, and combining the two objectives yields the best performance under the \textsc{Rouge} metrics. 
%We observe that though the relevance and stance auxiliary tasks both increase the \textsc{Rouge} and \textsc{BertScore}, less improvements can be attributed to the stance classification task. This is potentially due to the difference in performance between the two pretrained models we used to produce relevance stance and soft labels ($92\%$ vs. $75\%$).

To empirically verify whether the perspectives generated by our multi-task model are improved in terms of the relevance and stance correctness, we again use the two pretrained BERT classifiers to measure the percentage of generated summary with correct relevance and stance label. The results potentially suggest that by ``mimicing'' the predictions made by the two pretrained classifiers, our multi-task framework is able to generate summaries with higher quality along the two dimensions.    

%Similarly, we observe that the model benefits more from the relevance than the stance task.
% In this section, we present evaluation results of ur argument summarization system with auxiliary tasks. To guarantee the improvements are not induced by the stochastic nature of neural networks, we train each model for 3 different seeds and report the averaged results. The results are shown in Table \ref{tab:result}. The hyperparameters used for the second and third rows are $\alpha _1$ = 8 and $\alpha _2$ = 1.5, respectively. And the model with both auxiliary tasks use $\alpha _1$ = 50, $\alpha _2$ = 1.5. The tuning process of the hyperparameters is shown in Appendix \ref{sec:hyperparameters}.
% The F1 accuracy scores are reported for ROUGE and BERTscore. The Rel Score represents the likelihood of generated summary relevant to the query, and the Stance Accuracy shows the proportion of generated summaries with correct stance. 

% \siyi{Discuss the results here}
% The improvements shown in Rel Score and Stance Accuracy prove that both auxiliary signals significantly help the model to generate summaries with their specific objectives. Besides, they help the model generate a better summary in general as shown from the growth in ROUGE and BERTscore. (More discussion of the results with upcoming updated results...)

\subsection{Human Evaluations}
We randomly sampled 100 instances of abstracts with query from the test set, and ask two human raters to judge the quality of perspectives generated by the four systems. For the four summaries generated from an abstract by the different systems, we shuffle their order and ask the raters to rank each summary by the overall quality, with four criteria considered (1) \textit{Fluency} (2) \textit{Grammatical Correctness} (3) \textit{Faithfulness to the arguments offered in the original abstract} (4) \textit{Salience}. We allow ties among different summaries. We report their averaged ranks and the number of times a system is ranked first place in Table \ref{tab:human_eval}. The results are the averaged scores between the two annotators, and the level of agreement between them for this 4-class ranking task is $\kappa = 0.35$. 

For each summary, we ask the raters to annotate whether it (1) represents a relevant argument to the query (2) constitutes the correct stance as the target stance label. The kappa agreement between the two raters for these two tasks are 0.54 and 0.70, respectively.
We show the human evaluation results in Table~\ref{tab:human_eval}. 
We observe that while both the relevance and stance auxiliary tasks improve the quality of the generated perspective, combining the two auxiliary tasks does not guarantee a better summary quality.

\subsection{Analysis and Discussion} \label{sec:7}

\begin{table}
    \centering
    \begin{tabular}{|c|c|}
    \hline
         \textbf{Query} &\makecell {Should trump accept democrats’ \\ gov’t spending bill?} \\
         \hline
         \hline 
         \textbf{\textsc{Bart}}& \makecell{A shutdown is the best \\  deal he can get. } \\
         \hline
         \textbf{\textsc{+ Rel}} & \makecell{Trump should accept \\ the gop budget. } \\
         \hline 
         \hline
         \textbf{Gold} & \makecell{This deal is the most \\ achievable compromise. } \\
         
         \hline
    \end{tabular}
    \caption{An example where relevance auxiliary task helps the perspective summarization process}
    \label{tab:ablation}
\end{table}

\begin{table}
    \centering
    \begin{tabular}{|c|c|}
    \hline
         \textbf{Query} &\makecell {Is apple's iphone x technology\\ any good?} \\
         \hline
         \hline 
         \textbf{\textsc{Bart}}& \makecell{Apple's new iPhone X offers \\many great opportunities} \\
         \hline
         \textbf{\textsc{ + Stance}} & \makecell{Apple's new face-recognition \\technology raises many ethical \\ issues}  \\
         \hline 
         \hline
         \textbf{Gold} & \makecell{Apple's new Iphone X raises \\ many security concerns} \\
         
         \hline
    \end{tabular}
    \caption{An example where stance auxiliary task helps the perspective summarization process}
    \label{tab:ablation2}
\end{table}

% In this section we will provide examples of generated summaries and show how auxiliary tasks help with our target argument summarization task. 
The results on \textsc{Rouge}, \textsc{BertScore} and human evaluation suggest that the perspective summarization model learning benefits from both the relevance and stance tasks. However, we also observe that the vanilla BART present a strong baseline in both automatic and human evaluations.  

We list two typical cases where we observe the relevance and stance objectives improve the quality of the generated summary. 
For the query shown in Table~\ref{tab:ablation}, the BART model generates an out-of-context word ``shutdown'', which exists in the abstract, but is not applicable in the context provided by the query. The model with relevance objective, on the other hand, generates a perspective that is coherent to the context provided. 
For the query shown in Table~\ref{tab:ablation2}, the baseline BART model incorrectly produces a supporting perspective to the query, while the editorial or abstract presents the opposite stance. The model with the stance objective generates a perspective with a matching stance.

While we choose relevance and stance classification as the two auxiliary tasks in this case study, there exist many other candidate tasks that might be helpful in the setting. For instance, measuring the quality \cite{toledo-etal-2019-automatic}, or more specifically persuasiveness \cite{carlile-etal-2018-give} of the perspective might be two, amongst other, viable options. As our study assumes that the abstract is provided for each editorial, the overall performance of perspective summarization will likely drop, if we use model-generated abstract instead of ground truth as input.

\section{Related Work}

\subsection{Argumentation in News Editorials}
News editorials have been studied as a resource for studying many argumentation-related tasks. \citet{wilson-wiebe-2003-annotating, yu2003towards} use editorials for the study on sentiments and opinions. Later works \cite{reed-etal-2008-language, bal2009towards, chow-2016-argument} shift focus on the argumentation structure within editorials, and their persuasiveness effect \cite{al2016news, el-baff-etal-2020-analyzing}. A few other recent studies have explored argument quality \cite{el-baff-etal-2018-challenge} and generation \cite{el-baff-etal-2019-computational} when using editorials as a resource.

Our proposed dataset and study focus on the interplay between elements of the argumentation structure presented in editorial articles. Unlike previous work, we study these elements as the abstractive instead of extractive summary from the news editorials. 

\subsection{Argument Generation}
Most early efforts in argument generation, i.e. generating components in an argumentation structure, study rule-based synthesis methods based on argumentation theories 
\cite{reed1996architecture,zukerman-etal-2000-using}. With the recent progress in neural, sequence to sequence text generation methods \cite{sutskever2014sequence}, a few studies have adapted such techniques for end-to-end argument generation. 
\cite{wang-ling-2016-neural, hua-wang-2018-neural, hua-etal-2019-argument-generation}. 

The task of perspective generation in this work closely relates to argument conclusion generation \cite{alshomary-etal-2020-target}.
% , i.e. Generating a concise conclusion, or \emph{perspective} in an argument. 
Our study focuses on the setting where the target topic, or the \emph{query}, is given as input to the generation model. Due to the implicit nature of the \emph{perspectives} \cite{habernal2018argument}, one key challenge to the task is keep the semantics of the perspective generated \textit{truthful} to the abstract and editorial article. We approach this by measuring the compatibility of the perspective to the context along the dimensions of content salience \cite{bar-haim-etal-2020-arguments} and stance correctness \cite{bar-haim-etal-2017-stance}. Our multi-task generation approach conceptually resembles the work by \citet{guo-etal-2018-soft}, where multiple auxiliary tasks is employed to improve the quality of the generated summary. 
% \subsection{Stance Detection}

\section{Conclusion}
We present \datasetname\, an open-domain news editorial corpus that induces a number of argumentation-related tasks. The proposed dataset presents a few properties that are absent from existing datasets. First, the elements in the annotation structure are presented as abstraction over the text in editorial, as such elements usually exist implicitly in editorials. Second, as the pairs of editorials are aligned by topic, and exhibit opposing stance to each other, such structure allows for studies on cross-document argumentation structure. Third, the dataset allows for the study of multiple argumentation-related tasks together.

To demonstrate the power of having multiple related tasks in a single high-quality dataset, we study the problem of perspective summarization in a multi-task learning setting. Our analysis shows that modeling stance and relevance classification jointly with the summarization task improves the overall quality of the perspective generated. 

In future work, we hope to utilize the corpus to improve the multi-task framework for perspective summarization. As we set aside the problem of abstract generation in our case study, we would also like to identify the challenges and potential solution to the problem. We hope that \datasetname\ presents opportunities and challenges to future research in argumentation. 

% As the perspective  that contains argumentative data with a natural structure of perspectives, consists of {\em Query}, {\em Perspective}, {\em Abstract}, and {\em Article}. We claim that this structure of argumentative data induce many potential tasks, and provide an multi-task learning approach for one of the generation tasks. Our multi-task learning setting leverages the structure of our dataset by including auxiliary tasks using other instances in our dataset. We also show that this auxiliary task learning architecture helps our generation system produce better argument summarization, with control of the stance and relevance attributes of the generated perspective.

\section*{Ethical Considerations}

We collected data for \datasetname \ by automatically extracting data from \url{www.theperspective.com/perspectives}. The CEO of the website, Daniel Ravner, granted us permission to extract and use their data for academic research. We further annotated the data using crowd-workers. All crowd-workers were compensated by a fair wage determined by estimating the average completing time of each annotation task. Please refer to section \ref{sec:4_2} for more details.

The queries, abstracts, and perspectives in \datasetname \ are written by the professional writers of the website. The website aims at presenting the perspectives in each article without unnecessary subjective interpretation, but there is no guarantee that no subjectivity is involved in their content creation process.

\section*{Acknowledgments}
The authors would like to thank Daniel Ravner, the CEO of \url{www.theperspective.com}, for kindly granting access to data from the site for academic research.
This work was supported in part by a Focused Award from Google, and a gift from Tencent.

\bibliography{anthology,ccg,cited_10s,new}
\bibliographystyle{acl_natbib}

\newpage
\appendix

\section{Training Details} \label{sec:training_details}
\label{sec:appendix}

\subsection{Experiment Settings}

In this section, we describe our experiment settings in more details for reproducibility. We randomly split the \datasetname\ dataset into 70\%/10\%/20\% splits for training, validation, and testing, respectively. We train each system for 6 epochs using the same training set, and use the validation set to find the best $\alpha_1$ and $\alpha_2$. We report the test set results in Table \ref{tab:result}. All test set results are averaged results using three different random initilizations.  %Besides, for systems that incorporate auxiliary tasks, the first two epochs are trained only on BART without the auxiliary tasks. The intuition is that we want the system to first learn to generate better and stable summaries before the auxiliary tasks step in. 
 %we exclude the stance auxiliary loss if a query is an "open-ended" question and does not have stance labels.  
The approximate training time for a system trained with 6 epochs on a 12GB GPU is less than an hour. 

 For results shown in Table \ref{tab:result}, we use $\alpha_1 = 30$ in \textsc{+ Rel},  $\alpha_2 = 1$ in \textsc{+ Stance}, and $\alpha_1 =1, \alpha_2=1$ in \textsc{ + Rel \& Stance}, as they achieved the best results in the dev set during hyperperamater tuning \ref{sec:hyperparameters}.

\subsubsection{BART}
BART  pre-trained model has been proved effective on text generation, question answering, and summarization tasks \citep{lewis-etal-2020-bart}. Given the limited size of our dataset, we finetune on BART to transfer learn from the large amount of data it was pre-trained on. We use BART base with 6 encoder and decoder layers with hidden size of 768. We use AdamW with learning rate 3e-5 as our optimizer.

\subsubsection{BERT Relevance Classifier} \label{sec:rel_bert}
BERT pre-trained model has demonstrated its power in question answering, language inference, and text classification tasks \citep{devlin-etal-2019-bert}. We finetune BERT-mini on {\em PERSPECTRUM} dataset first on relevance classification task \citep{CKYCR19}, and then finetune it on our dataset, yielding an accuracy of 92\% on evaluation set (20\% of the data). The finetuned BERT model has 4 layers and hidden size of 256. 

\subsubsection{BERT Stance Classifier} \label{sec:stance_bert}
As the Relevance Classifier, we also finetune it on {\em PERSPECTRUM} dataset before training on our dataset.  However, we use a larger BERT model with 8 layers and hidden size of 768 since it is a slightly more difficult task than the relevance task. For the 30 queries (2\% of the whole corpus) that are labeled as open-ended questions and do not constitute a clear stance, we exclude them from the training of BERT stance classifier. Our classifier eventually achieves 75\% accuracy in the 20\% evaluation set.

\subsection{Hyperparameter Tuning} \label{sec:hyperparameters}

To control the degree we penalize our model using auxiliary loss, we introduce hyperparameters $\alpha$. We use the 10\% validation set to choose our best parameters. We tune $\alpha$ for different values from 0.1 to 50 to examine its affect on our model. We choose the best $\alpha$ according to their relevance and stance scores, and if there is a tie, we select the one with the higher ROUGE2 score. Table \ref{tab:rel_alpha}, Table \ref{tab:stance_alpha}, and Table \ref{tab:rel_stance_alpha} show the validation set results for tuning the BART+Rel, BART+Stance, and BART+Rel \& Stance systems, respectively. 

\begin{table}[h]
    \centering
    \begin{tabular}{c|cc}
    \hline
         $\alpha_1$ & Relevance Score \\
         \hline
         0.1 & 90.0 \\
       
         1 & 91.43 \\
       
         5 & 88.21\\
      
         15 & 92.5\\
    
         \textbf{30} &  \textbf{92.5} \\
         
         50 & 92.5 \\
         
         \hline
    \end{tabular}
    \caption{Tuning $\alpha_1$ for BART + Rel. Here we choose $\alpha_1=30$ since it has the highest ROUGE2 score.}
    \label{tab:rel_alpha}
\end{table}

\begin{table}[h]
    \centering
    \begin{tabular}{c|cc}
    \hline
         $\alpha_2$ & Stance Score \\
         \hline
         0.1 & 72.14 \\
       
         \textbf{1} & \textbf{72.50} \\
       
         5 & 64.64\\
      
         15 & 64.29\\
    
         30 &  60.36\\
         
         50 & 62.50 \\
         \hline
    \end{tabular}
    \caption{Tuning $\alpha_2$ for BART + Stance}
    \label{tab:stance_alpha}
\end{table}

\begin{table}[h]
    \centering
    \begin{tabular}{cc|ccc}
    \hline
        $\alpha_2$ & $\alpha_1$ & Relevance & Stance \\
         \hline
        0.1& 1 & 92.86 & 71.07 \\
       
        0.1& 5  & 88.21 & 63.93 \\
       
        0.1 & 50  & 92.5 &64.64\\
    
        \textbf{1} & \textbf{1} &  \textbf{93.93} &  68.93\\
        
        1 & 5 & 90.71 & \textbf{71.79} \\
        
        1 & 50 &93.57& 70.36 \\
         \hline
    \end{tabular}
    \caption{Tuning $\alpha_1$ and $\alpha_2$ for BART + Rel \& Stance. Here we choose ($\alpha_1=1$, $\alpha_2=1$) over ($\alpha_1=5$, $\alpha_2=1$) since it has a higher ROUGE2 score.}
    \label{tab:rel_stance_alpha}
\end{table}

\subsection{Measure of Agreement} \label{sec:agreement}
We use Cohen's and Fleiss kappa to measure the inter-rater agreement among annotators \cite{doi:fleiss-cohen}. We calculate Cohen's kappa agreement when there are only two raters, and Fleiss's kappa when there are more than two. 

Cohen's Kappa:
Let n be the number of instances to be labeled by A and B two raters. g is the number of distinct categories, and $f_{ij}$ denotes the frequency of the number of subjects with the ith categorical response for rater A and the jth categorical response for rater Y. The kappa agreement is then calculated as 
\begin{align*}
    p_0 &= \frac{1}{n}\sum_{i=1}^g f_{ii} \\ 
    p_e &= \frac{1}{n^2} \sum_{i=1}^g f_{i+} f_{+i} \\
    \kappa &= \frac{p_0-p_e}{1-p_e}
\end{align*}

where $f_{i+}$ is the total for the ith row $f_{+i}$ and is the total for the ith column in the frequency table.

Fleiss Kappa: Let N be the total number of subjects, let n be the number of ratings per subject, and let k be the number of categories into which assignments are made. Let $n_{ij}$ represent the number of raters who assigned the i-th subject to the j-th category. The kappa agreement is calculated as

\begin{align*}
    p_i &=  \frac{1}{n(n-1)} \sum_{j=1}^k n_{ij} (n_{ij}-1) \\
    \bar{P} &= \frac{1}{N} \sum_{i=1}^N p_i \\
    p_j &= \frac{1}{Nn} \sum_{i=1}^N n_{ij}\\
    P_e &= \sum_{j=1}^k p^2_j \\
    \kappa &= \frac{\bar{P}-P_e}{1-P_e}
\end{align*}

\section{Example Screenshots from ThePerspective Website and MTurk Annotation Interface} \label{sec: screenshots}

In this section, we show example screenshots from the website where we extract the data, \url{www.theperspective.com/perspectives}, and example screenshots of our data annotation process. 

For the three data annotation tasks using Mechanical Turk, stance annotation, implicit reference resolution, and topic annotation, we present the Turkers with definitions and instructions of the tasks that we require them to do, and 3-6 example questions with our expected answers. We ask them to read and comprehend our instructions before annotating, and use random control sets to filter out invalid annotations. More details can be found in the screenshots below.

\begin{figure*}[h]
    \centering
    \includegraphics[width=16cm]{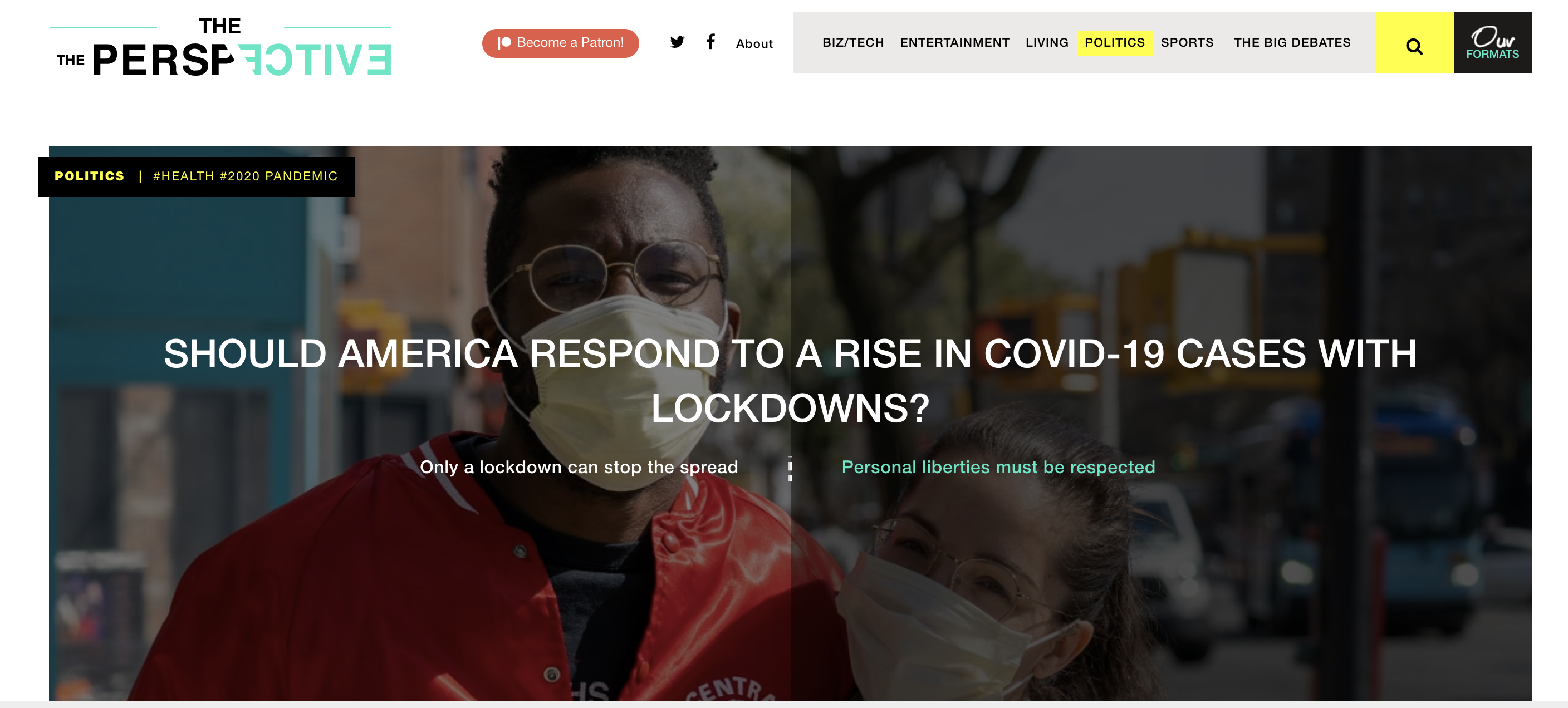}
    \caption{An example of a query and its two perspectives in ThePerspective website}
    \label{fig:persWebsite}
\end{figure*}

\begin{figure*}[h]
    \centering
    \includegraphics[width=16cm]{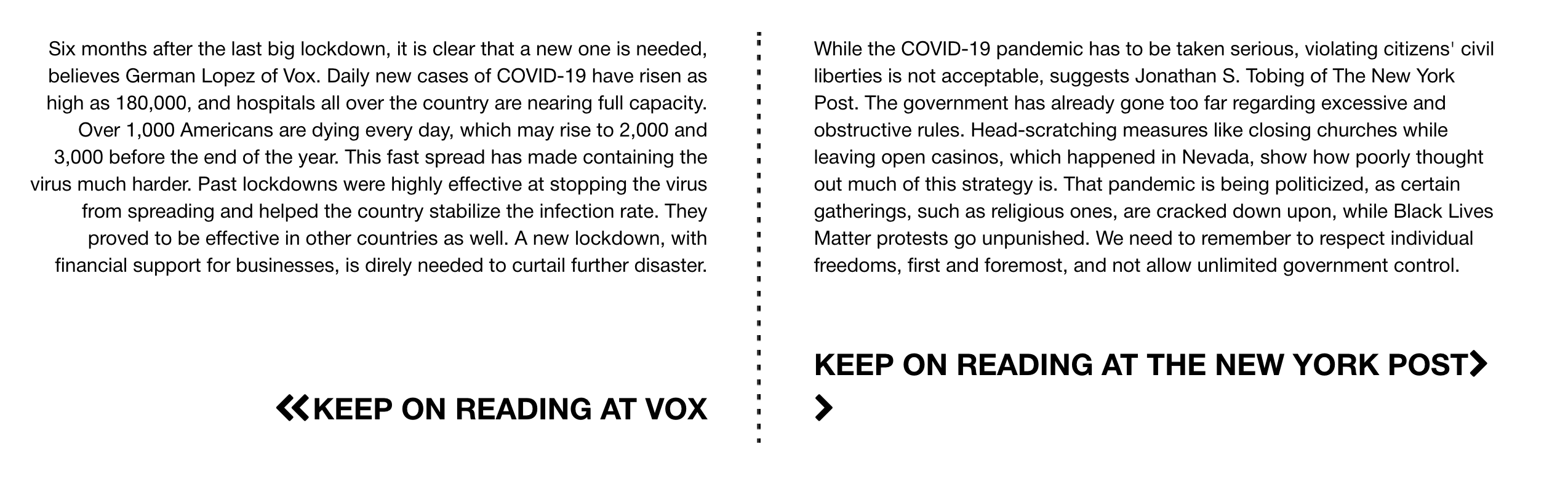}
    \caption{An exmaple of two abstracts and their links to news editorials in ThePerspective website}
    \label{fig:persWebsite2}
\end{figure*}

\clearpage

\begin{figure*}
    \centering
    \includegraphics[width=14cm]{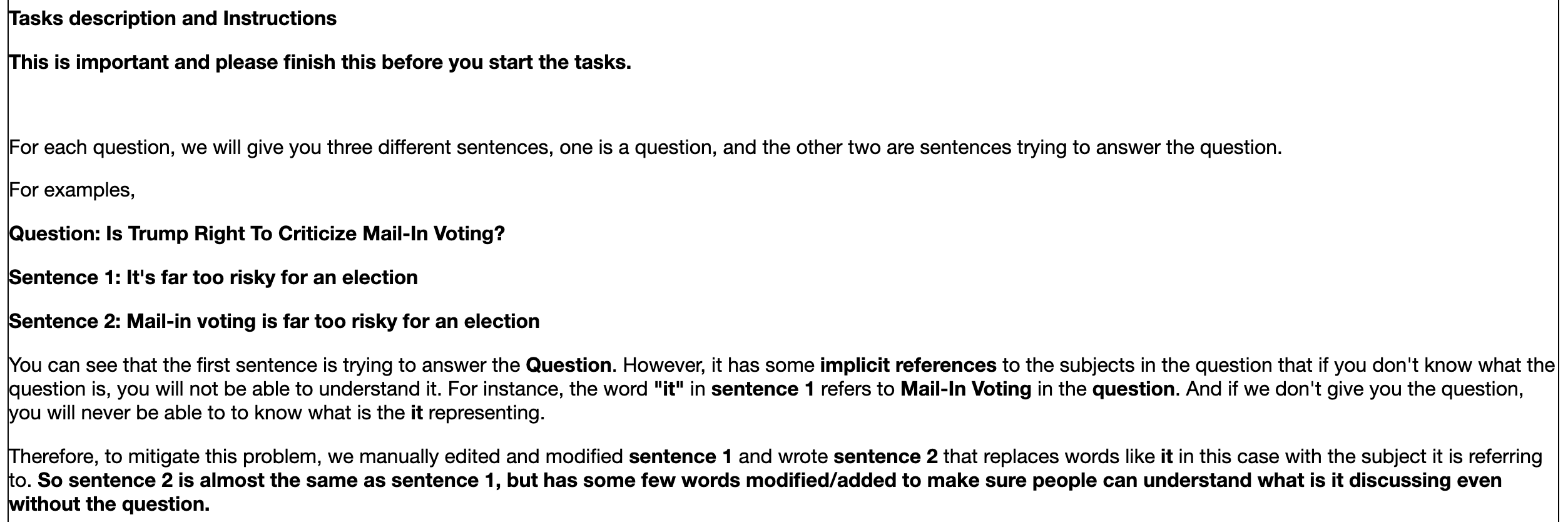}
    \caption{A screenshot of the MTurk annotation instruction for implicit reference resolution part1}
    \label{fig:annotation_ref1}
\end{figure*}

\begin{figure*}
    \centering
    \includegraphics[width=14cm]{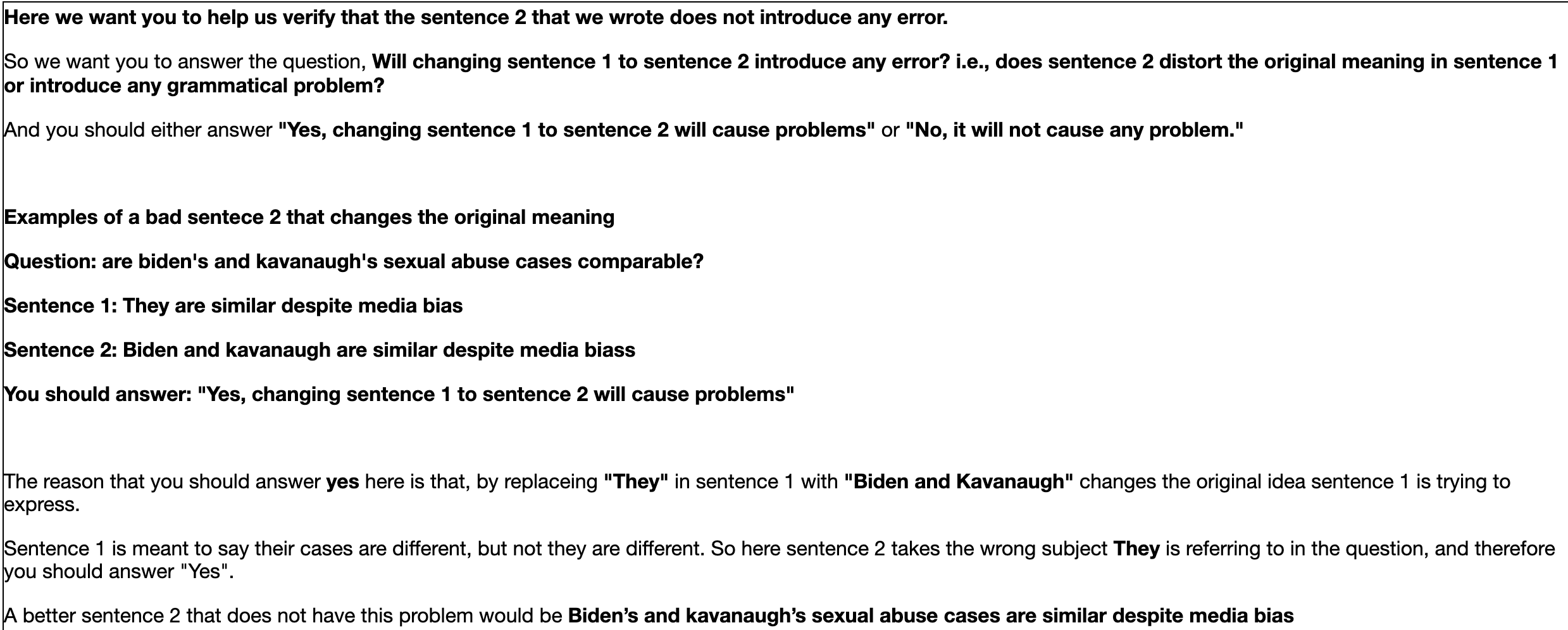}
    \caption{A screenshot of the MTurk annotation instruction for implicit reference resolution part2}
    \label{fig:annotation_ref2}
\end{figure*}

\begin{figure*}
    \centering
    \includegraphics[width=14cm]{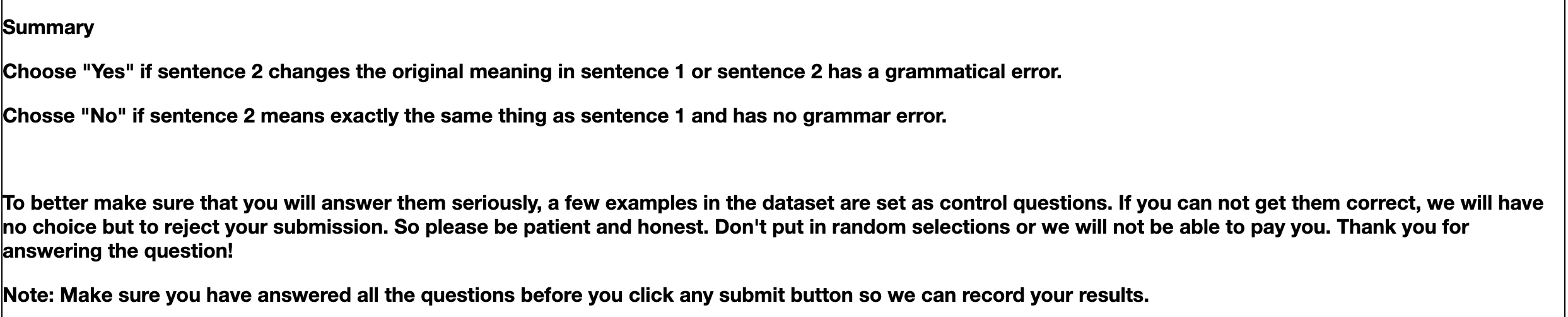}
    \caption{A screenshot of the MTurk annotation instruction for implicit reference resolution part3}
    \label{fig:annotation_ref3}
\end{figure*}

\begin{figure*}
    \centering
    \includegraphics[width=14cm]{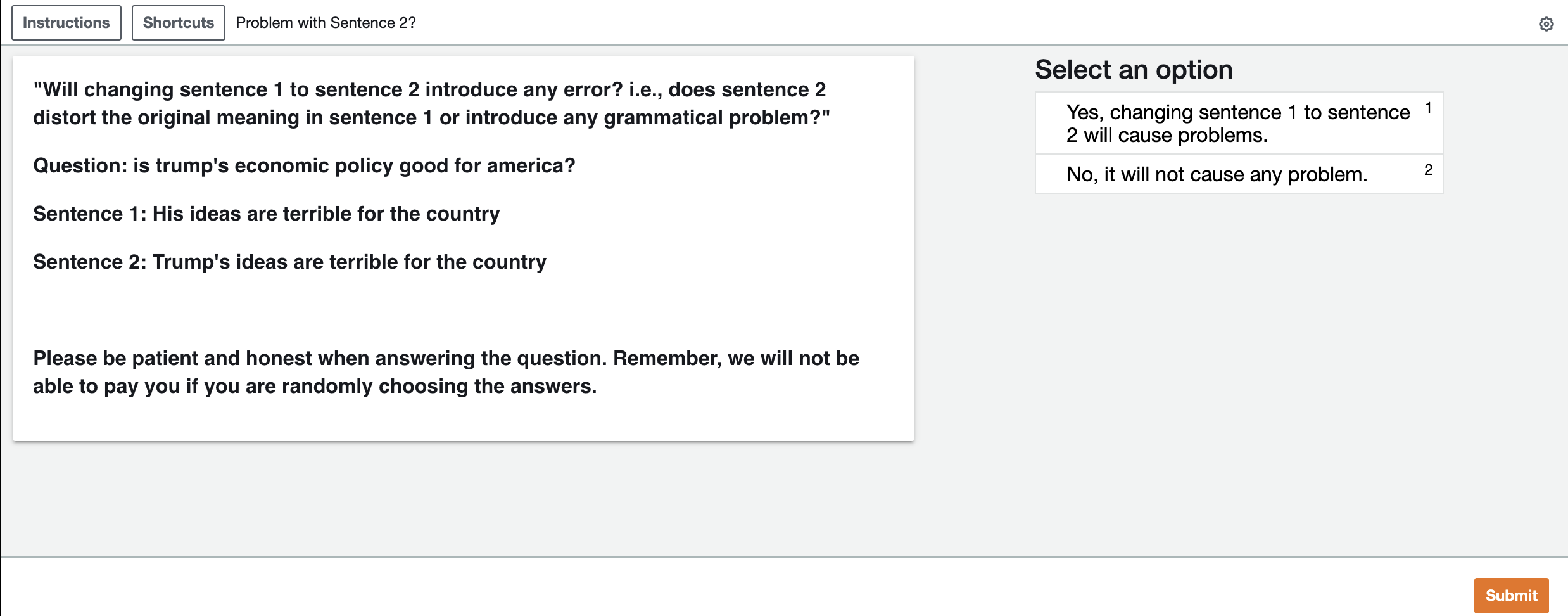}
    \caption{A screenshot of the MTurk annotation example for implicit reference resolution}
    \label{fig:annotation_ref4}
\end{figure*}

\clearpage

\begin{figure*}
    \centering
    \includegraphics[width=14cm]{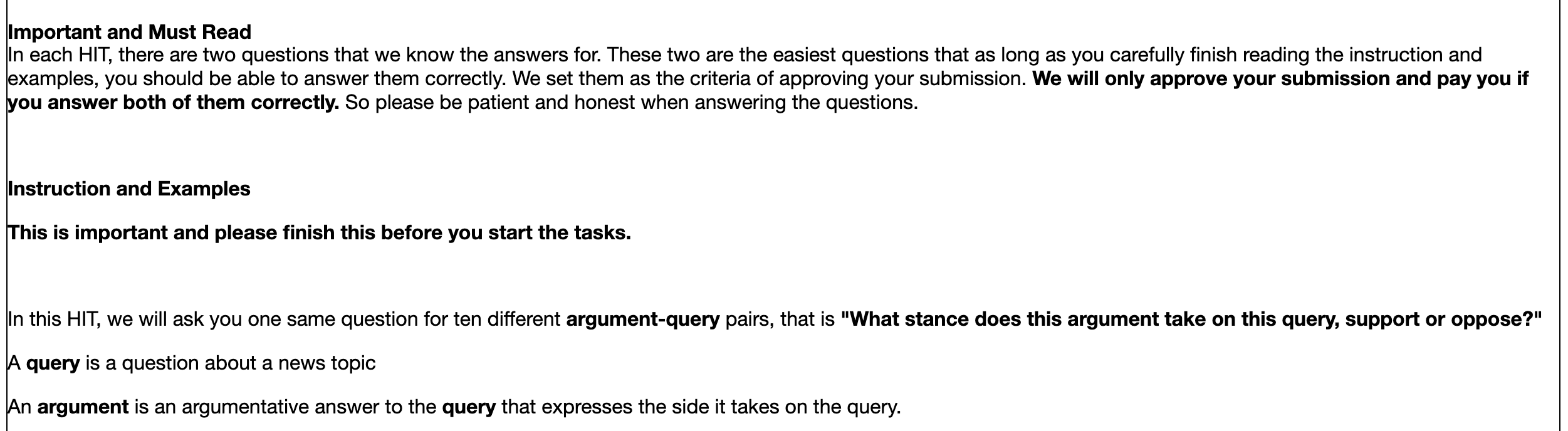}
    \caption{A screenshot of the MTurk annotation instruction for stance annotation part1}
    \label{fig:annotation_stance2}
\end{figure*}

\begin{figure*}
    \centering
    \includegraphics[width=14cm]{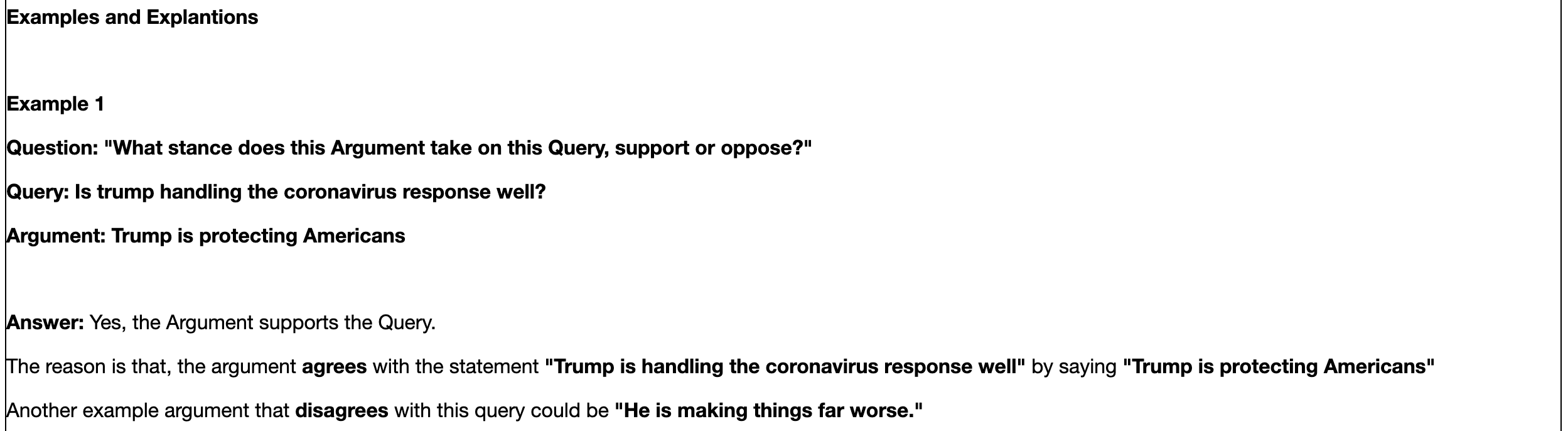}
    \caption{A screenshot of the MTurk annotation instruction for stance annotation part2}
    \label{fig:annotation_stance3}
\end{figure*}

\begin{figure*}
    \centering
    \includegraphics[width=14cm]{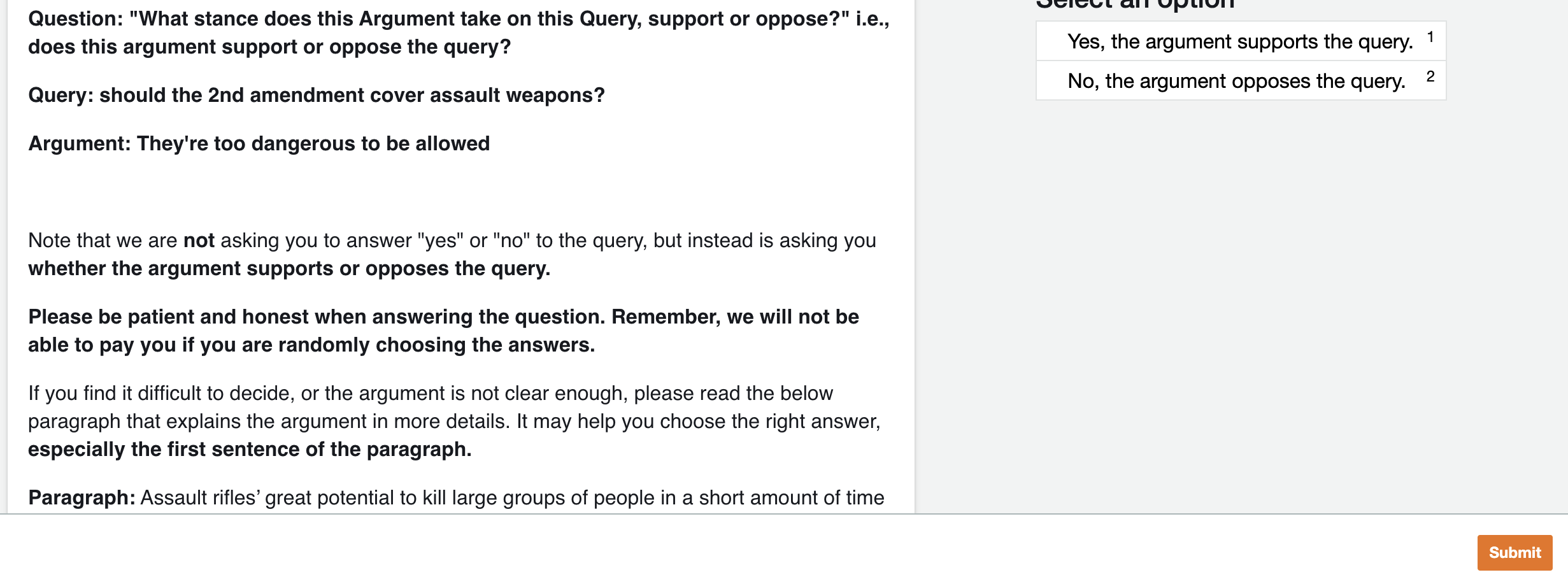}
    \caption{A screenshot of the MTurk annotation example for stance annotation. Note that there will be more sentences shown in the Paragraph line if the user scrolls down.}
    \label{fig:annotation_stance4}
\end{figure*}

\clearpage

\begin{figure*}
    \centering
    \includegraphics[width=13cm]{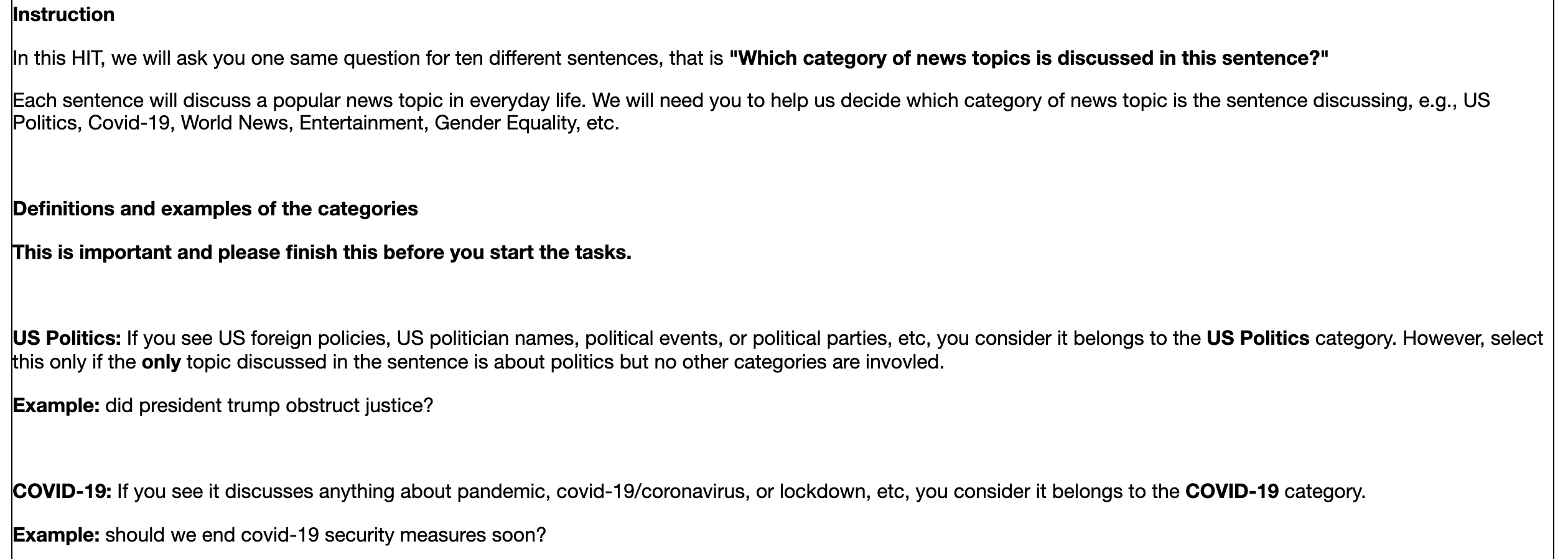}
    \caption{A screenshot of the MTurk annotation instruction for topic annotation part1}
    \label{fig:annotation_topic1}
\end{figure*}

\begin{figure*}
    \centering
    \includegraphics[width=13cm]{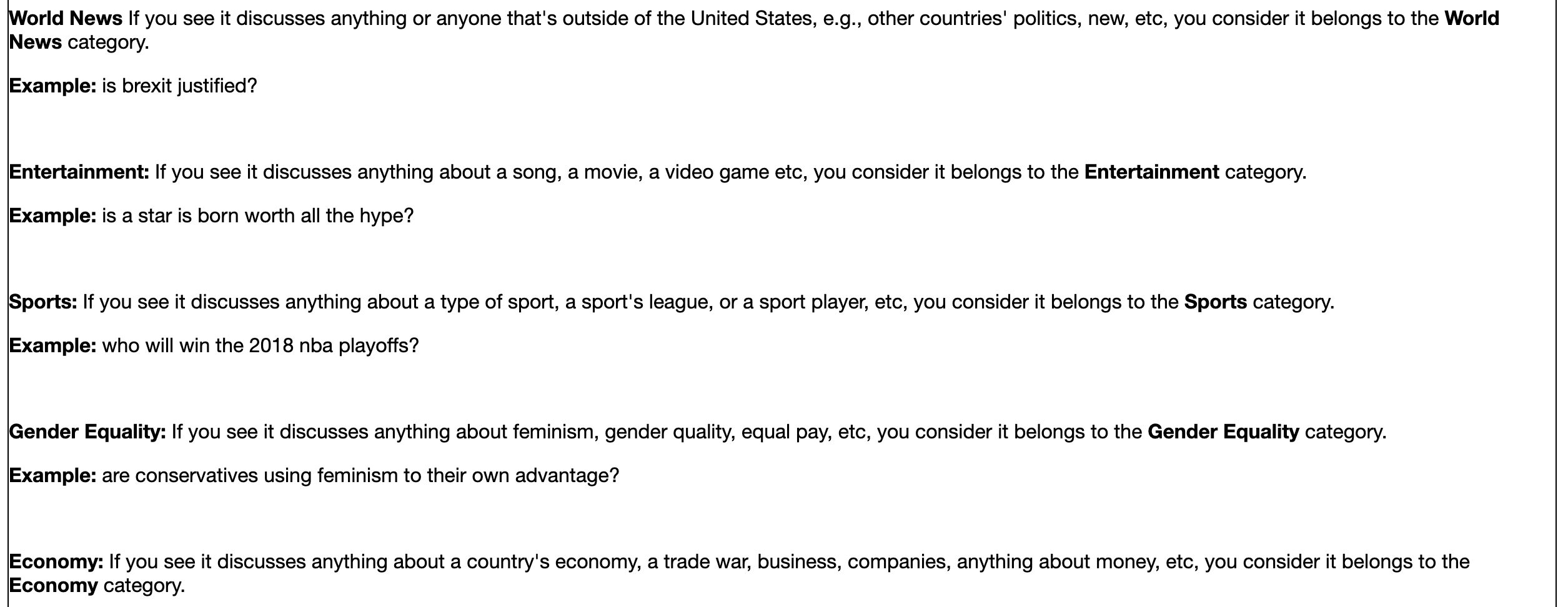}
    \caption{A screenshot of the MTurk annotation instruction for topic annotation part2}
    \label{fig:annotation_topic2}
\end{figure*}

\begin{figure*}
    \centering
    \includegraphics[width=13cm]{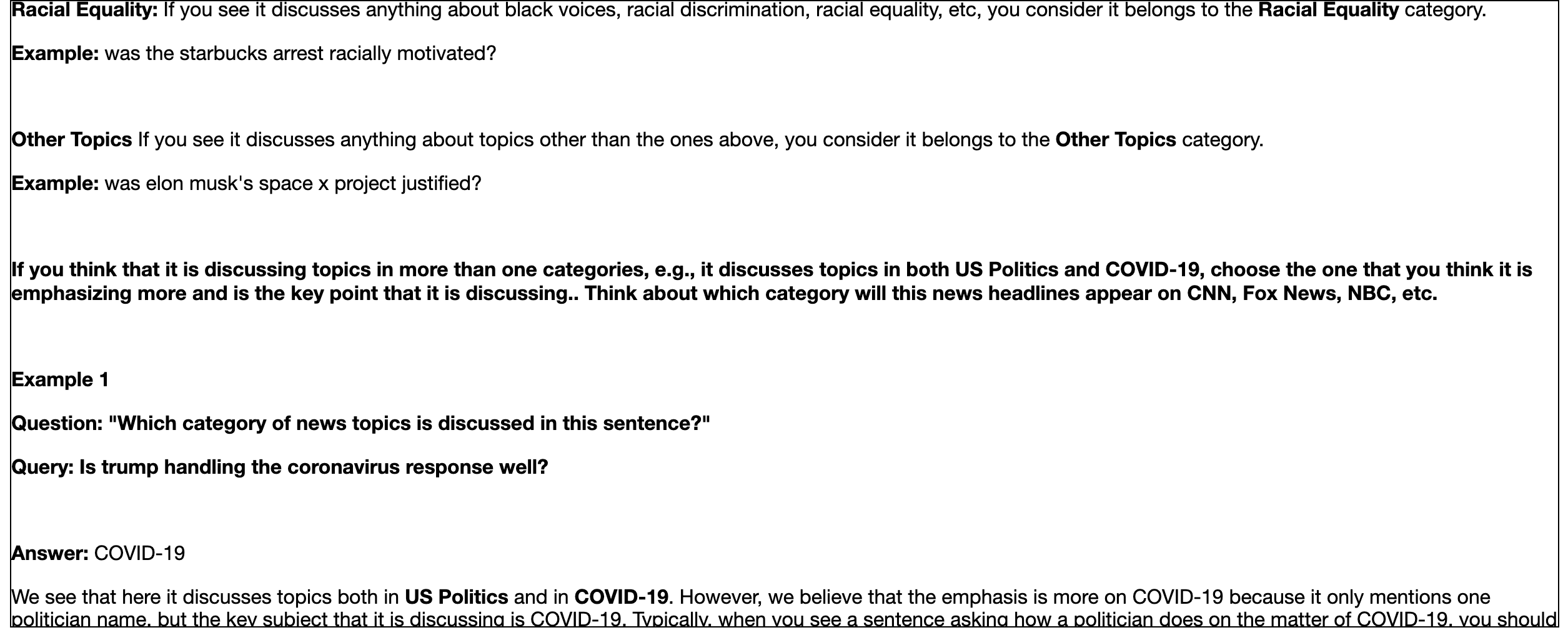}
    \caption{A screenshot of the MTurk annotation instruction for topic annotation part3}
    \label{fig:annotation_topic3}
\end{figure*}

\begin{figure*}
    \centering
    \includegraphics[width=13cm]{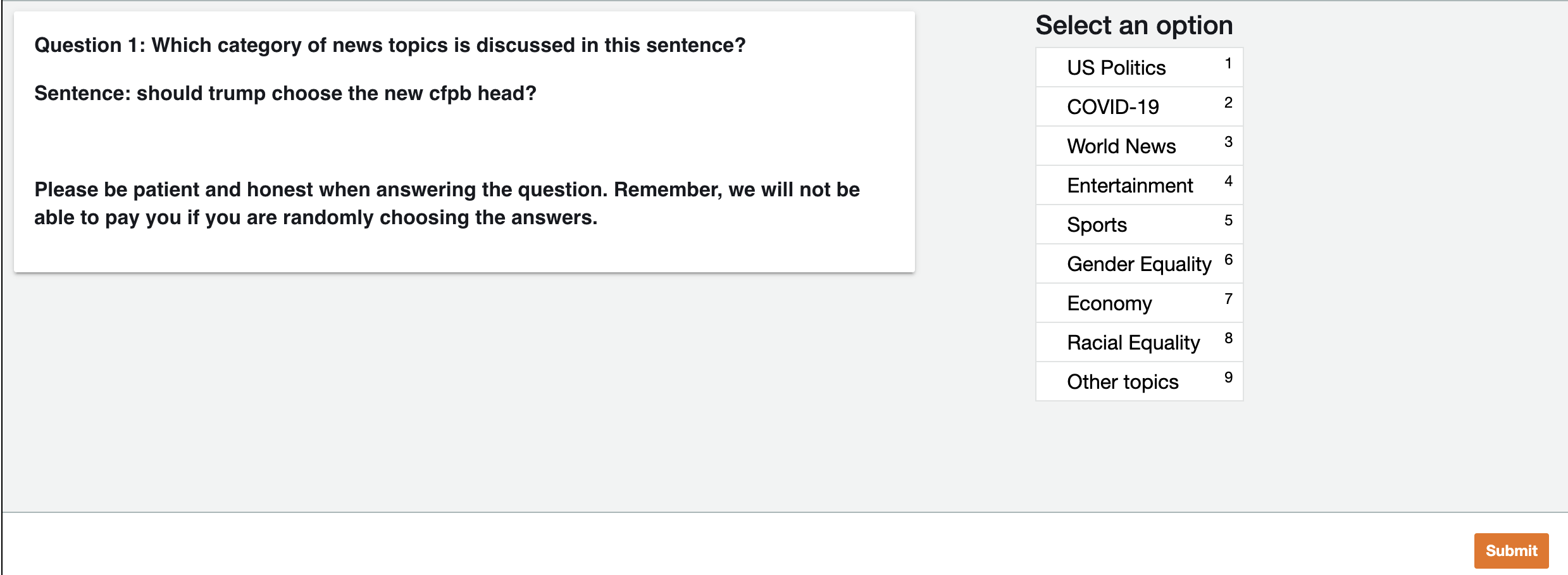}
    \caption{A screenshot of the MTurk annotation example for topic annotation}
    \label{fig:annotation_topic4}
\end{figure*}

\end{document}